%% file: main.tex
\documentclass[10pt,twocolumn,letterpaper]{article}

\usepackage[pagenumbers]{cvpr} 
\usepackage{multirow}
\usepackage{booktabs}
\usepackage{arydshln}
\usepackage{float}
\usepackage{url}
\input{preamble}

\definecolor{cvprblue}{rgb}{0.21,0.49,0.74}
\usepackage[pagebackref,breaklinks,colorlinks,citecolor=cvprblue]{hyperref}

\def\paperID{1757} 
\def\confName{CVPR}
\def\confYear{2025}

\title{Q-Eval-100K: \underline{E}valuating \underline{V}isual Quality and \underline{A}lignment \underline{L}evel \\ for Text-to-Vision Content}

\author{Zicheng Zhang$^{1}$ \quad Tengchuan Kou$^{1}$\quad Shushi Wang$^{1}$ \quad Chunyi Li$^{1}$ \\  Wei Sun$^{1}$ \quad Wei Wang$^2$ \quad  Xiaoyu Li$^2$\quad Zongyu Wang$^2$ \quad
Xuezhi Cao$^2$ \\  Xiongkuo Min$^{1}$\quad  \stepcounter{footnote}Xiaohong Liu$^{1}\thanks{~Corresponding author.}$\quad  Guangtao Zhai$^{1}$ \\
$^{1}$Shanghai Jiao Tong University \quad $^{2}$Meituan
}

\begin{document}

\maketitle

\begin{abstract}
Evaluating text-to-vision content hinges on two crucial aspects: visual quality and alignment. While significant progress has been made in developing objective models to assess these dimensions, the performance of such models heavily relies on the scale and quality of human annotations. According to \textbf{Scaling Law}, increasing the number of human-labeled instances follows a predictable pattern that enhances the performance of evaluation models. Therefore, we introduce a comprehensive dataset designed to \underline{E}valuate \underline{V}isual quality and \underline{A}lignment \underline{L}evel for text-to-vision content (\textbf{Q-EVAL-100K}), featuring the largest collection of human-labeled Mean Opinion Scores (MOS) for the mentioned two aspects.
The \textbf{Q-EVAL-100K} dataset encompasses both text-to-image and text-to-video models, with 960K human annotations specifically focused on visual quality and alignment for 100K instances (60K images and 40K videos). Leveraging this dataset with context prompt, we propose \textbf{Q-Eval-Score}, a unified model capable of evaluating both visual quality and alignment with special improvements for handling long-text prompt alignment. 
Experimental results indicate that the proposed \textbf{Q-Eval-Score} achieves superior performance on both visual quality and alignment, with strong generalization capabilities across other benchmarks. These findings highlight the significant value of the \textbf{Q-EVAL-100K} dataset. Data and codes will be available at \url{https://github.com/zzc-1998/Q-Eval}.
\end{abstract}

\begin{figure}[h]
    \centering
    \includegraphics[width=.85\linewidth]{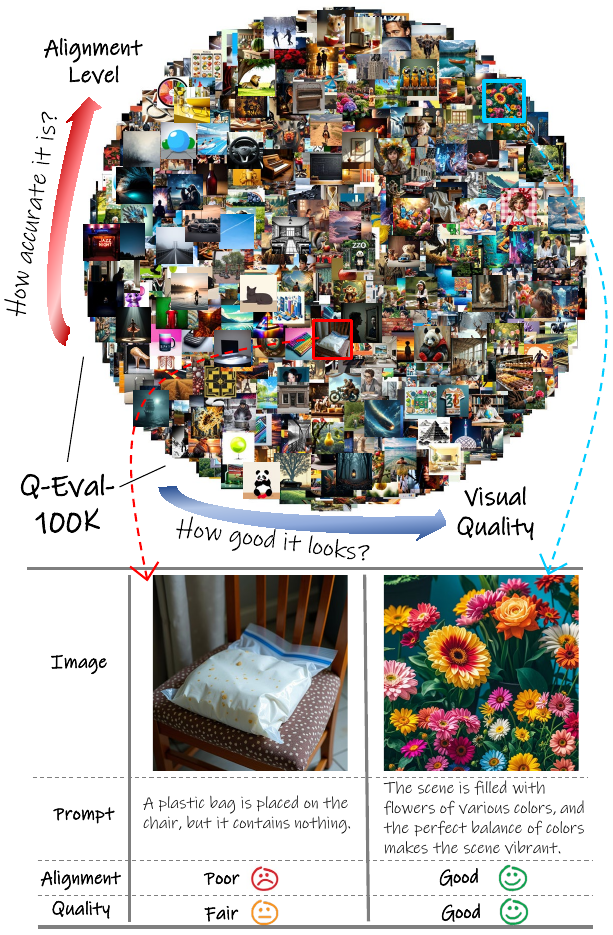}
    \caption{Illustration of the unified evaluation dimensions of \textbf{Q-Eval-100K}. We focus on \textbf{visual quality} (including all factors that may impact the viewing experience) and \textbf{alignment level}, which measures the accuracy of the generated content to the prompt.  }
    \label{fig:spotlight}
    \vspace{-0.4cm}
\end{figure}

\section{Introduction}


With the rapid advancement of generative AI, millions of text-to-image and text-to-video content are being generated daily across various platforms~\cite{everypixel2024,techreport2024}, applied in industrial production or directly used by consumers. However, due to current technological limitations, text-to-vision content often falls short of being perfect upon generation and cannot be immediately deployed~\cite{agiqa3k,vqascore,Q-Refine}, which usually requires expert evaluation, editing, and fine-tuning. As a result, numerous efforts have been made to develop automated methods for evaluating text-to-vision content, aiming to control the quality of generation and guide the necessary optimizations effectively~\cite{pickapic,imagereward,hpdv2,ku2023viescore,kim2023prometheus,zhang2023gpt,cho2023davidsonian,vqascore,q-instruct,qbench,wu2024qalign,zhang2024qboost}. Through extensive theoretical and experimental analysis, the evaluation of text-to-vision content can be primarily divided into two dimensions~\cite{imagereward,hps,abench,zhang2024survey,liu2024ntire}: \textbf{Visual Quality} (the perceived quality of the visual content, which can be simply understood as \textit{how good it looks}) and \textbf{Alignment} level (the consistency between text and vision, which can be interpreted as \textit{how accurate the generation is}).

\begin{table}[!t]
    \centering
    \renewcommand\arraystretch{1.1}
    \caption{A brief comparison of the latest text-to-vision evaluation datasets (\textit{I. for image, V. for video}). For \textbf{Annotation Type}, \textbf{SBS} (\textit{side-by-side}) and \textbf{MOS} (\textit{mean-opinion-score}) refer to selecting the preferred instance from a pair of instances and assigning an average absolute score to a single instance respectively. For \textbf{Rating Concerns}, \textit{Overall} indicates assigning scores from a holistic perspective while \checkmark denotes assigning separate scores to quality or alignment. For \textbf{Number}, \textit{Ins.} and \textit{Ann.} stand for the number of instances and human annotations respectively.}
    \vspace{-6pt}
   \resizebox{\linewidth}{!}{\begin{tabular}{l|c|c|c|cc|c}
    \toprule
   \multirow{2}{*}{\textbf{Dataset}} & \multirow{2}{*}{\textbf{Year}} & \multirow{2}{*}{\textbf{Content}} & {\textbf{Annotation Type}} & \multicolumn{2}{c|}{\textbf{Evaluation Concern}} & \textbf{Number}  \\ \cdashline{4-7}
   &&&\textit{(Single/Pair)} & \textit{Quality} & \textit{Alignment} & \textit{Ins.}/\textit{Ann.}\\ 
   \hline
    Pick-a-pic~\cite{pickapic} & 2023 & I. & SBS & \multicolumn{2}{c|}{\textit{Overall}} & 1M/500K\\
    ImageReward~\cite{imagereward} & 2023 & I. & SBS & \multicolumn{2}{c|}{\textit{Overall}} & 68k/137k \\
    HPDv2~\cite{hpdv2} & 2023 & I. & SBS & \multicolumn{2}{c|}{\textit{Overall}} & 430k/645K \\ \hdashline
    AGIQA-3k~\cite{agiqa3k} & 2023 & I. & MOS & \checkmark & \checkmark & 3K/81K \\
    AIGCIQA2023~\cite{wang2023aigciqa2023} & 2023 & I. & MOS & \checkmark & \checkmark & 2K/17K \\
   PKU-AIGIQA-4k~\cite{yuan2024pku} & 2024 & I. & MOS & \checkmark & \checkmark & 4K/84K \\ 
   AIGIQA-20k~\cite{li2024aigiqa} & 2024 & I. & MOS & \multicolumn{2}{c|}{\textit{Overall}} & 20K/420K \\
   RichHF~\cite{richhumanfeedback} & 2024 & I. & MOS & \checkmark & \checkmark & 18K/54K \\
   VideoFeedback~\cite{he2024videoscore} & 2024 & V. & MOS & \checkmark & \checkmark & 37.6K/37.6K \\
   T2VQA-DB~\cite{kou2024subjective} & 2024 & V. & MOS & \multicolumn{2}{c|}{\textit{Overall}} & 10K/27K\\
   GenAI-Bench~\cite{li2024genaibench} & 2024 & I.\&V. & 1-5 Likert Scale & \multicolumn{2}{c|}{\textit{Overall}} & 9.6K/9.6K \\
    \textbf{Q-Eval-100K (Ours)} & 2024 & I.\&V. & MOS & \checkmark & \checkmark & 100K/960K \\
    \bottomrule
\end{tabular}}
\vspace{-10pt}
    \label{tab:comparison}
\end{table}

To meet the need for evaluation, many text-to-vision evaluation datasets have been proposed, along with corresponding evaluation algorithms~\cite{liu2024ntire,chen2024gaia,chen2023natural,he2024videoscore,huang2024vbench,agiqa3k,lmmpcqa,wang2023aigciqa2023,yuan2024pku,li2024aigiqa, richhumanfeedback,he2024videoscore,li2024genaibench,kou2024subjective,qbenchvideo,qbench+}. However, these efforts face the following significant limitations:
1) \textbf{The key evaluation dimensions for text-to-vision content are often not systematically captured.} Some datasets propose too many dimensions, adding unnecessary complexity to the evaluation process. The practical applicability of these dimensions can be narrow or lead to redundancy.  
2) \textbf{Most text-to-vision evaluation datasets fail to disentangle visual quality and alignment.} These datasets either focus solely on alignment or visual quality, or merge both dimensions into a single score, leading to results that are often incomplete and ambiguous, making it challenging to address specific evaluation needs.
3) \textbf{The scale of these datasets remains insufficient.}With the rise of Large Multimodal Models (LMMs), which have demonstrated strong capabilities in visual and textual understanding, researchers are increasingly leveraging them for text-to-vision evaluation. However, current dataset sizes remain inadequate to fully unlock the potential of LMM-based models~\cite{zhang2024survey}, conceivably restricting their applicability and generalization in real-world scenarios.


To address these challenges, we present \textbf{Q-Eval-100K}, which, to the best of our knowledge, is the largest text-to-vision evaluation dataset with Mean Opinion Scores (MOSs), comprising 100K instances (including 60K generated images \& 40K generated videos). A brief comparison of \textbf{Q-Eval-100K} and previous text-to-vision evaluation datasets is illustrated in Table~\ref{tab:comparison}. We manually gather prompts from existing benchmarks and create diverse prompts that focus on three key aspects: \textit{entity generation}, \textit{entity attribute generation}, and \textit{interaction capability}. The instances in \textbf{Q-Eval-100K} are then generated from a diverse range of generative models, both open-source and closed-source, to ensure high diversity. We implement a rigorous, scientifically grounded subjective evaluation process using a \textbf{Sample \& Scrutinize} strategy, focusing on both visual quality and alignment level for each of the 100K instances, yielding a total of high-quality 960K human annotations.

Building on the proposed \textbf{Q-Eval-100K}, we propose \textbf{Q-Eval-Score}, a unified evaluation framework capable of assessing both visual quality and alignment, providing separate scores for each dimension. We first adapt \textbf{Q-Eval-100K} into a Supervised Fine-Tuning (SFT) dataset optimized for injecting knowledge into LMMs. Scores are transformed into \textit{adjective-based ratings}, then reformulated within a well-guided \textit{context-prompt format}. Specifically, for visual quality, we guide the model to identify positive and negative visual impacts, evaluate the intensity of these impacts, and make a balanced judgment. For alignment, we guide the model to perceive the overall situation, examine details, and reach a balanced judgment. The fine-tuning process is then supervised by a combined CE and MSE loss.
During inference, the final score is computed as a weighted average based on \textit{the probability of each rating token.} Notably, in handling long-prompt alignment, we observed that direct alignment assessment often yields low scores due to oversimplification. To address this, we propose a \textbf{Vague-to-Specific} strategy, where a long prompt is converted into a vague version retaining only core information and multiple prompts with specific details. These prompts are evaluated separately and the alignment scores are combined to the final score.
Our contributions can be summarized as follows:
\begin{itemize}
    \item We present \textbf{Q-Eval-100K, the largest text-to-vision evaluation dataset with MOSs}, comprising 100K instances from various generative models. We employ a scientifically grounded evaluation methodology, using a \textbf{Sample \& Scrutinize} strategy to collect 960K human annotations focusing on visual quality and alignment.
    \item We propose \textbf{Q-Eval-Score}, a unified evaluation framework capable of independently assessing visual quality and alignment, providing separate scores for each dimension. Specifically, we adapt \textbf{Q-Eval-100K} into an SFT dataset with adjective-based ratings in a structured context-prompt format for enhancing the visual quality and alignment evaluation capabilities of LMMs.
    \item To improve alignment evaluation for long prompts, we introduce a \textbf{Vague-to-Specific} strategy, which separates prompts into core and detailed variants, yielding a more accurate alignment score through weighted averaging.
\end{itemize}

\section{Related Works}

\subsection{Benchmarks for Text-to-Vision Evaluation}
Early text-to-vision benchmarks largely depend on multimodal datasets labeled with captions~\cite{coco_caption,ok-vqa,hu2023tifa}. However, with increasing recognition of human feedback's value~\cite{rlhf,zhang2024survey,huang2023t2i}, many benchmarks begin to employ human annotations. Common annotation methods include SBS (side-by-side) and MOS (mean opinion score)~\cite{zerman2018relation,perez2019pairwise}. SBS requires selecting a preferred instance from a pair, while MOS assigns a score to a single instance. SBS is generally easier for human subjects and more precise, but MOS is more versatile and broadly applicable to various situations~\cite{leveque2021comparative,barratt2018note}.

Text-to-vision evaluation dimensions~\cite{liu2024ntire,abench} can be categorized into visual quality and alignment. While some benchmarks~\cite{chen2024gaia,chen2023natural,he2024videoscore,huang2024vbench} treat aspects like naturalness, aesthetics, and temporal consistency as distinct dimensions, we view these as components of visual quality since they collectively influence the quality of experience (QoE) for viewers.
Early benchmarks~\cite{agiqa3k,wang2023aigciqa2023,yuan2024pku,li2024aigiqa} for generated images comprehensively address both visual quality and alignment for evaluation. RichHF~\cite{richhumanfeedback} enhances these evaluations by incorporating subjective scores, heatmaps, and misalignment tokens. For video, VideoFeedback~\cite{he2024videoscore} introduces five dimensions of quality and alignment, while T2VQA-DB~\cite{kou2024subjective} focuses primarily on visual quality. Further, GenAI-Bench~\cite{li2024genaibench} evaluates alignment for both generated images and generated videos.
The proposed \textbf{Q-Eval-100K} dataset offers a unified text-to-vision evaluation framework, significantly increasing dataset scale and diversity, making it distinct from prior benchmarks.

\subsection{Metric for Text-to-Vision Evaluation}
Previous evaluation methods separately focus on either visual quality or alignment. Perceptual methods assess the visual quality of generated content, utilizing traditional scores like IS~\cite{inception}, FID~\cite{fid}, and LPIPS~\cite{lpips} with pre-trained neural networks. Recently, data-driven models~\cite{agiqa3k} trained on specialized datasets have further advanced perceptual score prediction. Additionally, methods such as CLIP-IQA~\cite{clipiqa} and Q-Align~\cite{wu2024qalign} leverage text prompts to enhance perceptual alignment.
Alignment methods, integrating both text and vision modalities, initially use CLIPScore~\cite{hessel2021clipscore} due to its ease of application. To address more complex prompts, some approaches~\cite{pickapic,imagereward,hpdv2} incorporate human feedback to improve evaluation accuracy. Given the powerful interpretative capabilities of LMMs, recent work~\cite{ku2023viescore,kim2023prometheus,zhang2023gpt,cho2023davidsonian,vqascore} has begun to apply these models to alignment assessments.
Most existing models evaluate either perceptual quality or alignment exclusively. The proposed \textbf{Q-Eval-Score} addresses this gap by offering decoupled scores for both perceptual quality and alignment.

\begin{figure}
    \centering
    \includegraphics[width=0.82\linewidth]{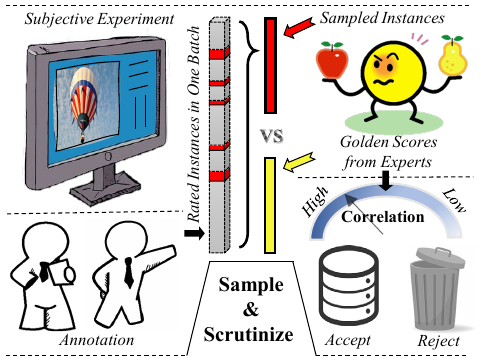}
    \caption{Illustration of the \textbf{Sample and Scrutinize} quality control strategy for annotations in \textbf{Q-Eval-100K}. We randomly select a sample of 5K instances from the full dataset, which are then reviewed by experts to establish golden scores. A batch of annotations is approved only if the scores of the sampled instances show a high correlation with these expert-assigned golden scores. }
    \label{fig:sample}
    \vspace{-12pt}
\end{figure}

\section{Q-Eval-100K Construction}

\subsection{Basic Principles}
\begin{figure*}
    \centering
    \includegraphics[width=\linewidth]{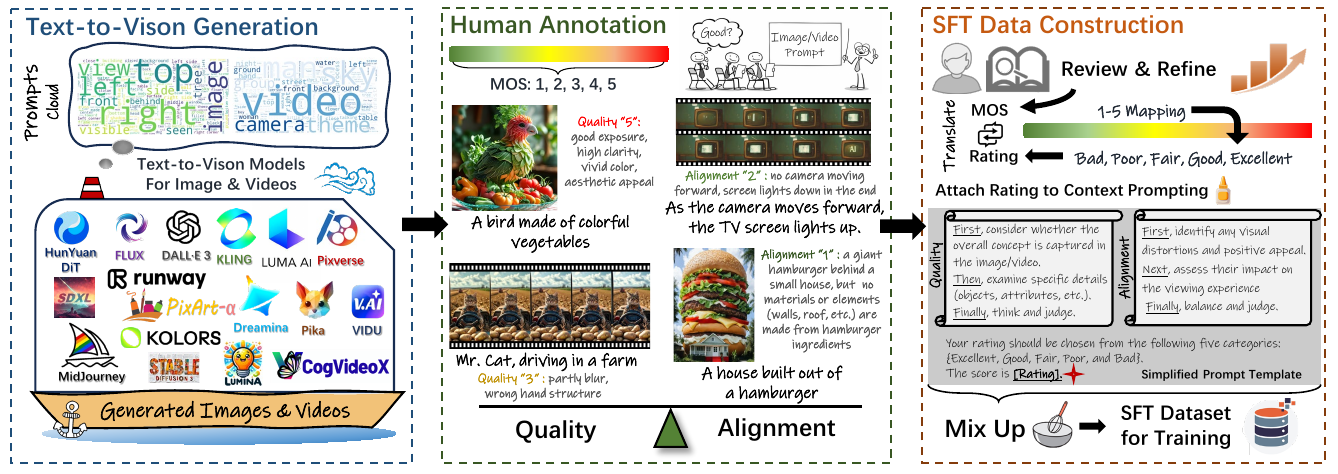}
    \caption{Overview of the \textbf{Q-Eval-100K} construction process. We design a wide range of prompts and employ various text-to-vision models to generate diverse content. Subjective evaluations are then conducted to rate the visual quality and alignment of these generated instances. The resulting scores form the SFT dataset, which can help inject corresponding knowledge into LMMs. }
    \label{fig:dataset}
\end{figure*}

The construction process of \textbf{Q-Eval-100K} is illustrated in Fig. \ref{fig:dataset}. We follow these guiding principles: 1) \textbf{Ensuring diversity} in generated content by collecting a wide range of prompts and using multiple generative models; 2) \textbf{Ensuring annotation quality} through carefully designed experimental settings and standards to achieve the highest accuracy; 3) \textbf{Ensuring effective learning}, through adapting the data for LMM suitability by transforming both visual quality and alignment scores into a context-aware SFT dataset.

\subsection{Sources Collection}

\textbf{Prompt Designing.} The prompt design focuses on three main aspects: \textbf{Entity Generation}, \textbf{Entity Attribute Generation}, and \textbf{Interaction Capability}. 1) Entity generation targets the primary entities (\textit{people, objects, etc.}) to be generated. 2) Entity attribute generation emphasizes the attributes (\textit{clothing, color, material, etc.}) of the entities.  3) Interaction capability focuses on the interactions between the generated entities and other entities or the background, such as their spatial relationships and actions. Following the outlines mentioned above, we manually create a portion of the prompts and extract some from existing datasets~\cite{li2024genaibench,onoe2024docci}.
\\
\textbf{Generation Models.} We utilize multiple popular text-to-image and text-to-video models to ensure diversity, which include FLUX~\cite{flux1_github},  Lumina-T2X~\cite{gao2024lumin-t2x}, PixArt~\cite{chen2023pixart}, Stable Diffusion 3~\cite{stablediffusion3}, Stable Diffusion XL~\cite{podell2023sdxl}, DALL·E 3~\cite{dalle3_openai}, Wanx~\cite{alibaba2024tongyiwanxiang}, Midjourney~\cite{midjourney}, Hunyuan-DiT~\cite{li2024hunyuan}, Kolors~\cite{kolors}, ERNIE-ViLG~\cite{feng2023ernie}, 
CogVideoX~\cite{yang2024cogvideox}, Runway GEN-2~\cite{germanidis2023gen2}, Runway GEN-3~\cite{runway2024gen3alpha}, Latte~\cite{ma2024latte}, Kling~\cite{kling}, Dreamina~\cite{dreamina}, Luma~\cite{luma_dream_machine}, PixVerse~\cite{pixverse_ai}, Pika~\cite{pika_ai}, Stable Video Diffusion~\cite{blattmann2023stable}, Vidu~\cite{vidu_ai}.

\subsection{Subjective Experiment}
Given the large scale of \textbf{Q-Eval-100K}, we develop rigorous experimental protocols to ensure the highest possible accuracy in our annotations. To facilitate this, we establish a well-controlled indoor experimental environment, where more than 200 human subjects are recruited to participate in the annotation. To ensure the accuracy of annotations is not compromised by individual cognitive differences or annotator fatigue, we propose a \textbf{Sample \& Scrutinize} data control strategy as shown in Fig.~\ref{fig:sample}. The strategy includes two steps:
1) First, we randomly sample 5,000 instances from the entire dataset. We then organize experts with rich experience to discuss and score these instances, leading to the establishment of golden scores for both visual quality and alignment. These golden scores remain hidden for all subsequent experimental subjects.
2) Next, we provide human annotators with comprehensive training before they begin the annotation process. After each batch, we gather the scores for instances that have golden scores and compare them with these golden scores, calculating the correlation values (SRCC-rank similarity). Only batches with an SRCC above 0.8 are accepted, otherwise, they are rejected.

Additionally, we split \textbf{Q-Eval-100K} into training and testing sets in an 80:20 ratio. Each instance in the training set has at least three annotations, while each instance in the testing set has a minimum of twelve annotations to ensure accuracy. This process results in a total of over 960K annotations, calculated as follows:
80K (\textit{training instances}) x 2 (\textit{visual quality \& alignment}) x 3 (\textit{minimum annotation number}) + 20K (\textit{training instances}) x 2 (\textit{visual quality \& alignment}) x 12 (\textit{minimum annotation number}) = 960K annotations.
Finally, we calculate the average of the multiple annotations to derive the score for each instance.



\begin{figure*}[htbp]
    \centering
    \begin{minipage}{0.47\textwidth}  
        \centering
        \includegraphics[width=\linewidth]{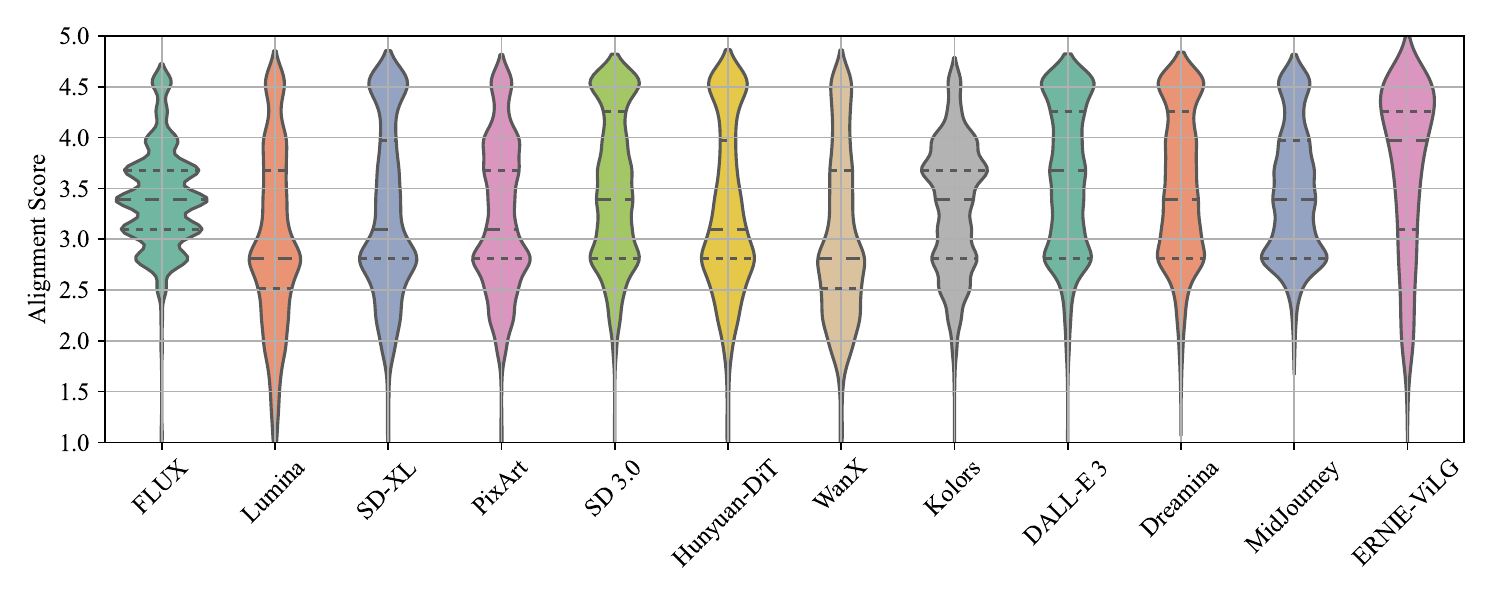}
        \subcaption{Image Alignment} \label{fig:img_alignment}
    \end{minipage}
    \begin{minipage}{0.47\textwidth}  
        \centering
        \includegraphics[width=\linewidth]{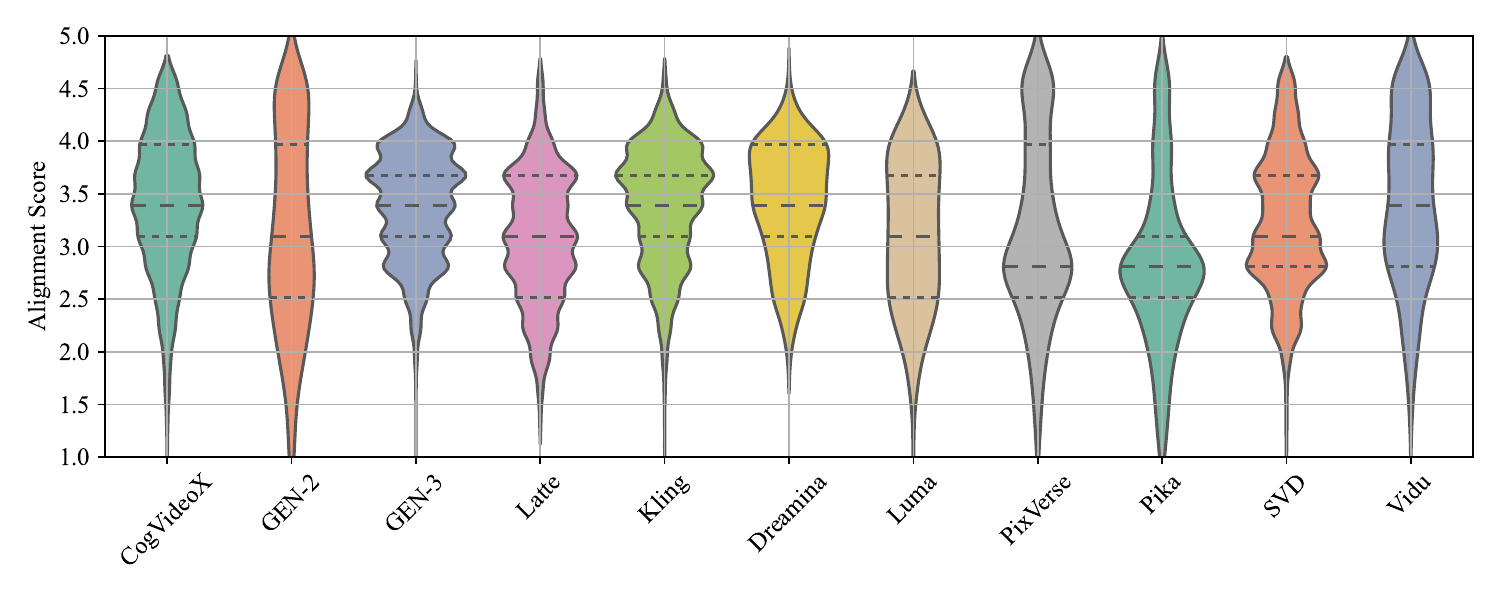}
        \subcaption{Video Alignment} \label{fig:vid_alignment}
    \end{minipage}
    \begin{minipage}{0.47\textwidth}
        \centering
        \includegraphics[width=\linewidth]{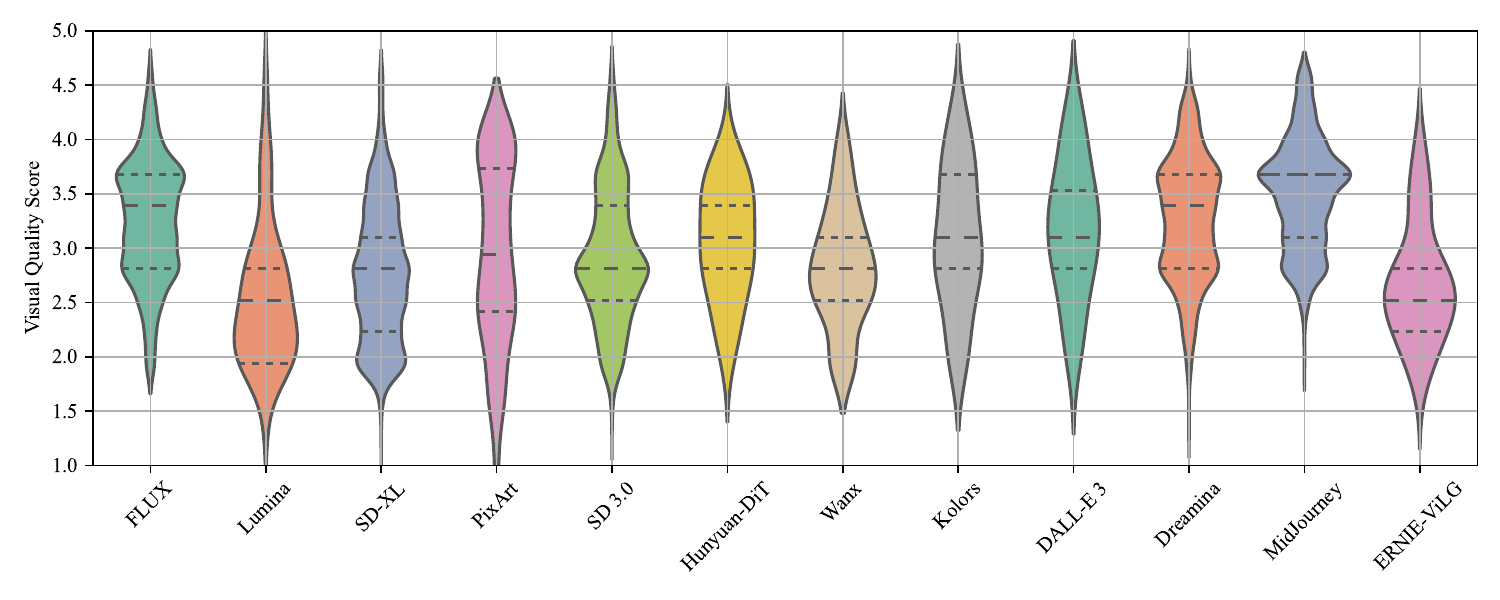}
        \subcaption{Image Visual Quality} \label{fig:img_quality}
    \end{minipage}
    \begin{minipage}{0.47\textwidth}
        \centering
        \includegraphics[width=\linewidth]{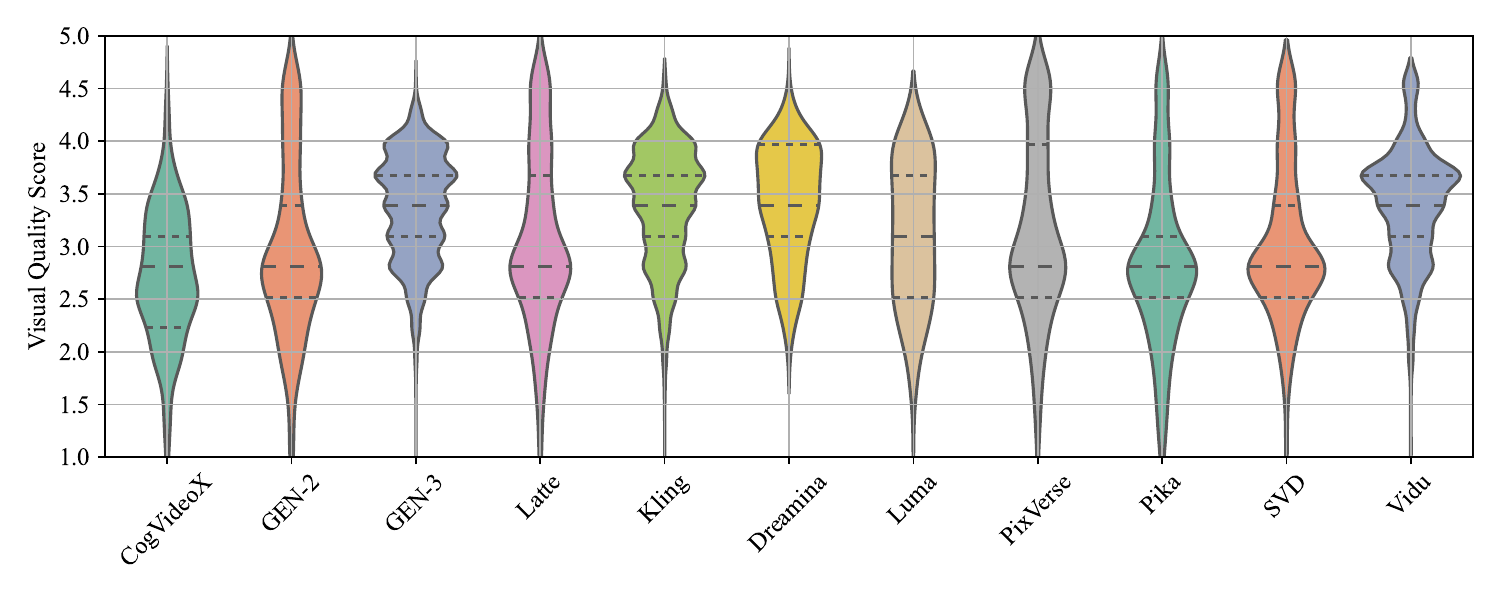}
        \subcaption{Video Visual Quality} \label{fig:vid_quality}
    \end{minipage}
    \caption{MOS distributions for the visual quality and alignment of generated images and videos in the \textbf{Q-Eval-100K} dataset respectively. }
    \label{fig:mos_distribution}
    \vspace{-6pt}
\end{figure*}

\subsection{Statistical Analysis}

The distributions of MOSs for visual quality and alignment are exhibited in Fig.~\ref{fig:mos_distribution} respectively,  which reveal several key insights. In general, \textbf{there are substantial differences among generation models in both visual quality and alignment}, with their distributions displaying significant inconsistencies, indicating varied performance across different generation prompts. 1) For image alignment, distributions are generally skewed higher, suggesting that models perform relatively well in aligning images, though multiple peaks in the 4-5 and 2-3 score ranges indicate some fluctuation in performance. 2) In video alignment, model performance varies more markedly, with most scores concentrated between 2 and 4, highlighting the need for improvement in alignment in video generation. 3) Visual quality for images scores noticeably lower than image alignment, indicating that generation models perform significantly worse in visual quality. Furthermore, the wider distribution spread in image visual quality suggests greater variance and instability across models. 4) Similarly, video visual quality scores are lower than alignment scores, highlighting a consistent underperformance in visual quality. Interestingly, models such as Kling, Dreamina, Luma, PixVerse, and Pika exhibit similar distributions for alignment and visual quality, indicating consistent capability across both aspects. However, this consistency is not observed across all models.

Overall, the findings above highlight a notable disparity in that visual quality generally falls behind alignment. This gap likely stems from the current emphasis on alignment optimization, which is also relatively easier to improve, whereas visual quality has received less focus. This analysis underscores the importance of \textbf{Q-Eval-100K} as a comprehensive benchmark for evaluating both dimensions.

\section{Q-Eval-Score}

\subsection{Unified Pipeline for Decoupled Evaluation}
Although the evaluation of visual quality and alignment are two relatively independent tasks, we leverage the adaptability and extensive prior knowledge of LMMs to propose a unified model, \textbf{Q-Eval-Score}, that addresses both visual quality and alignment level evaluation within a single framework. Specifically, we convert the human-labeled MOS from the \textbf{Q-Eval-100K} dataset for both visual quality and alignment levels into a fixed-prompt format, creating a mixed SFT dataset. We then fine-tune the LMM, enabling it to evaluate both visual quality and alignment levels.

\begin{figure*}
    \centering
    \includegraphics[width=0.95\linewidth]{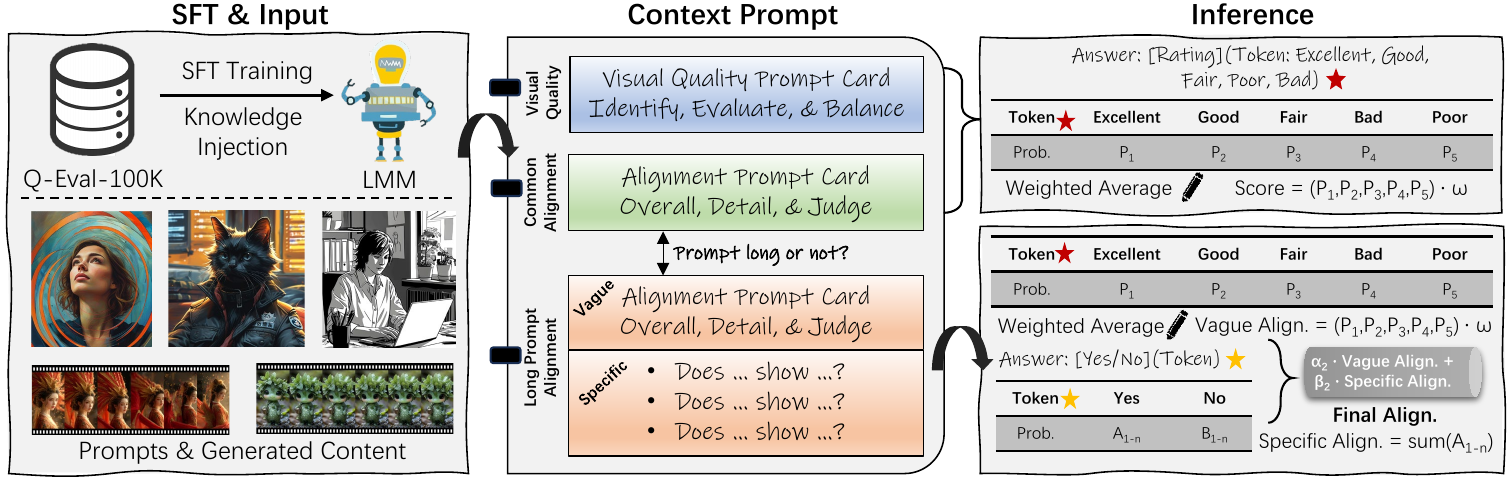}
    \caption{The pipeline of the proposed \textbf{Q-Eval-Score} model involves multiple stages. First, the \textbf{Q-Eval-100K} SFT dataset is used to train the LMM on visual quality and alignment knowledge. Then, context prompts are applied to guide the LMM towards generating more detailed and accurate outputs. Finally, the rating token probabilities are converted into predicted scores. Additionally, long prompt alignment is achieved through a \textbf{Vague-to-Specific} strategy to further refine the model's responses.  }
    \label{fig:q_eval_score}
\end{figure*}

\subsection{How to Teach LMMs to Evaluate}
\subsubsection{Context Prompt}
In previous work using prompts with LMMs for evaluation, the questions are often straightforward and simple, such as \textit{`Can you evaluate the quality of the image?'} (Q-Align~\cite{wu2024qalign}) or \textit{`Does this figure show [Prompt]? Please answer yes or no.'} (VQAScore~\cite{vqascore}). However, this simplicity may lead to confusion for the model, as the prompts may not be specific enough to guide a more detailed or accurate evaluation.

Inspired by the chain-of-thought (CoT~\cite{cot}) concept and given that humans undergo a reasoning process when evaluating visual quality and alignment, we propose a \textbf{Context-Prompt} format to construct our SFT dataset. For the visual quality task, the human evaluation process can be summarized as first identifying both positive and negative quality factors, then measuring these factors, and finally weighing them to reach a conclusion. Based on this process, we design the following prompt structure:

\noindent \textbf{Context Prompt for Visual Quality}

\noindent \textit{\# User: Suppose you are an expert in evaluating the visual quality of AI-generated image/video. \underline{First}, identify any visual distortions and positive visual appeal regarding low-level features and aesthetics. \underline{Next}, assess the severity of distortions and their impact on the viewing experience, noting whether they are subtle or distracting, and evaluate how the positive features enhance the visual appeal, considering their strength and contribution to the overall aesthetics. \underline{Finally}, balance the identified distortions against the positive aspects and give your rating on the visual quality.}
\textit{Your rating should be chosen from the following five categories: [Excellent, Good, Fair, Poor, and Bad]. For this image/video [Image/Frames], the text prompt is [Prompt].}

\noindent \textit{\# Answer: [Rating] (Excellent, Good, Fair, Poor, Bad).}

For the alignment task, the human evaluation process involves observing whether the overall content generally aligns with the text, followed by a more detailed comparison, and finally a comprehensive evaluation for conclusion:


\noindent \textbf{Context Prompt for Visual Quality}

\noindent \textit{\# User: Suppose you are an expert in evaluating alignment between the text prompt and the AI-generated image/video. \underline{Begin} by considering whether the overall concept of the prompt is captured in the image/video. \underline{Then}, examine the specific details, such as the presence of key objects, their attributes, and relationships. Check if the visual content accurately reflects these aspects. \underline{Finally}, give your alignment rating considering both overall and detailed accuracy. }
\textit{Your rating should be chosen from the following five categories: [Excellent, Good, Fair, Poor, and Bad]. For this image/video [Image/Frames], the text prompt is [Prompt]. }

\noindent \textit{\# Answer: [Rating] (Excellent, Good, Fair, Poor, Bad).}

\subsubsection{`Translating' MOS into Ratings}
It is well established that discrete adjective ratings are easier for LMMs to interpret compared to numerical scores~\cite{wu2024qalign,zhang2024qboost}. Since MOS in \textbf{Q-Eval-100K} is labeled in absolute terms, we can easily map MOS to the corresponding rating:
\begin{equation}
\begin{aligned}
 & R(s) \! \! = \!\! r_i \!\! \quad \! \! \text{if} \! \! \quad \! \! m \! + \! \frac{i\!-\!1}{5} \!\! \times \!\! (M \!\! - \!\!  m) \! < \! s \! \leq \! m \! + \! \frac{i}{5} \! \! \times \! \! (M \!\! -  \!\!m), \\
& \{r_i \mid_{i = 1 \sim 5} \} \! = \! \{Bad, Poor, Fair, Good, Excellent\}, 
\end{aligned}
\end{equation}
where $m = 1$ and $M = 5$ (score range bound of \textbf{Q-Eval-100K}), $R(s)$ indicates the mapped rating of MOS value $s$.

\subsection{Model Architecture}
Using the constructed SFT dataset with the question-answer pairs as described, we select Qwen-VL~\cite{Qwen-VL} as the LMM model (\textit{Qwen2-VL-7B-Instruct}) for training, which has demonstrated strong visual understanding capabilities for both images and videos. For video processing, each video is converted into a sequence of images at a rate of one frame per second, which is then fed into the model as the input. The scoring computation method is detailed as follows.
For the rating token, we first calculate the model output probabilities \( p_j \) for each of the five rating terms \( \{\textit{Excellent}, \textit{Good}, \textit{Fair}, \textit{Poor}, \textit{Bad}\} \), where \( j \in \{1, 2, 3, 4, 5\} \). Then we define the final predicted rating \( \hat{r} \) as the weighted average of these probabilities:
\begin{equation}
\hat{r} = \sum_{j=1}^5 p_j \cdot w_j,
\end{equation}
where \( w_j \) is the numerical weight assigned to each rating (e.g., \( w_j = \{1,0.75,0.5,0.25,0\} \) for \textit{Excellent} to \textit{Bad}).

\subsection{Loss Function}
The loss function for the model consists of two parts: Cross-Entropy (CE) Loss and Mean Squared Error (MSE) Loss. The CE Loss can assist the LMM in learning the general question-answer format and necessary knowledge. Meanwhile, the MSE Loss refines the score prediction accuracy. The CE Loss for question-answer pairs is defined as:
\begin{equation}
\mathcal{L}_{CE} = -\sum_{i=1}^N y_i \cdot \log(p_i),
\end{equation}
where \( y_i \) is the one-hot encoded vector representing the true label for instance \( i \), and \( p_i \) is the predicted probability vector for the answer tokens. The MSE Loss can then be given by:
\begin{equation}
\mathcal{L}_{MSE} = \left( \hat{r} - r_{\text{MOS}} \right)^2,
\end{equation}
where \( \hat{r} \) and \( r_{\text{MOS}} \) represent the predicted scores and the MOS labels repectively.
The total loss \( \mathcal{L} \) is a weighted sum of the CE Loss and MSE Loss:
\begin{equation}
\mathcal{L} = \alpha_1 \cdot \mathcal{L}_{CE} + \beta_1 \cdot \mathcal{L}_{MSE},
\end{equation}
where \( \alpha_1 \) and \( \beta_1 \) (\textit{default set as 1 \& 1}) are weight parameters controlling the contribution of each loss term.

\subsection{Handling Long Prompt Alignment}
During inference, we observe that the trained LMM tends to undervalue alignment when handling long prompts (more than 25 words). This is partly because long prompts are underrepresented in the training data, leading to insufficient training. More importantly, the LMM acts as a strict evaluator, often penalizing significant points for inconsistencies that may seem minor to humans. These small discrepancies occur more frequently with long prompts.
To manage this issue, we propose a \textbf{Vague-to-Specific} strategy. We use an additional LLM (QwenLM~\cite{qwen}) to summarize long prompts, retaining only the core features while filtering out details, producing a concise \textbf{Vague Prompt}. Then, we split the long prompt into \textbf{Specific Prompts} (\textit{no more than three}), each maintaining full details but avoiding redundancy:
\begin{equation}
    ({P}_{v}, \{{P}_{s_1}, \cdots, {P}_{s_n}\}) = \mathcal{VS}(P_{Long}),
\end{equation}
where ${P}_{v}$ represents the \textbf{Vague Prompt}, $\{{P}_{S_1}, \cdots, {P}_{S_n}\}$ stands for the set of \textbf{Specific Prompts}, $\mathcal{VS}(\cdot)$ indicates the prompt split function, and $P_{Long}$ is the original long prompt.
For the \textbf{Vague Prompt}, we calculate alignment in the usual way. 
However, directly asking for consistency with the \textbf{Specific Prompts} is not appropriate since each one addresses only part of the vision content.
Drawing inspiration from the VQAScore~\cite{vqascore} approach, we modify the question to a softer format, such as \textit{`Does the image/video show [Prompt]?'} to evaluate alignment (measuring as the logit probability of answering \textit{`Yes'}) for each \textbf{Specific Prompts}. Finally, we combine the results from both the \textbf{Vague Prompt} and \textbf{Specific Prompt} using a weighted approach to calculate the final alignment score:
\begin{equation}
    \mathcal{A}_{f} = \alpha_2 \mathcal{A}_{v} + \beta_2 \left( \frac{1}{n} \sum_{i=1}^{n} \mathcal{A}_{s_i} \right),
\end{equation}
where $\mathcal{A}_{f}$, $\mathcal{A}_{v}$, and $\mathcal{A}_{s_i}$ are the alignment scores for the final evaluation, vague prompt, and $i$-th specific prompt. $\alpha_2$ and $\beta_2$ (\textit{0.5 \& 0.5 as default}) are weight parameters.

\begin{figure}
    \centering
    \includegraphics[width=.93\linewidth]{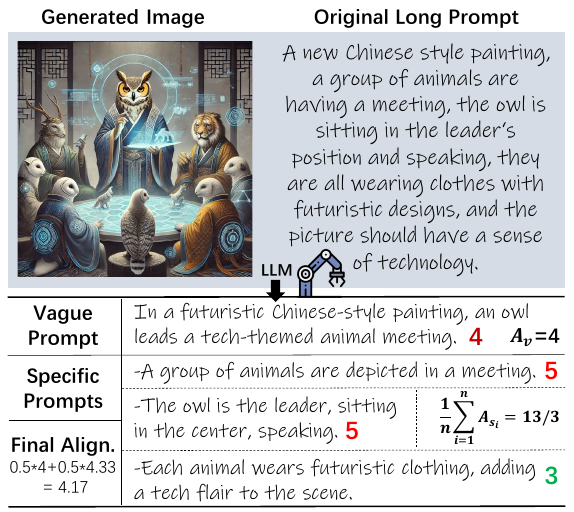}
    \caption{An example of the \textbf{Vague-to-Specific} strategy. The original long prompt is divided by the LLM (QwenLM~\cite{qwen}) into a \textbf{Vague Prompt} and several \textbf{Specific Prompts}. The alignment score is first calculated separately for each part, then combined using weighted averaging to form the final score. }
    \label{fig:vague}
    \vspace{-12pt}
\end{figure}

\begin{table}[!t]\small
    \centering
    \renewcommand\arraystretch{1.08}
    \setlength{\tabcolsep}{4pt}
    \caption{Performance comparison on the visual quality aspect of \textbf{Q-Eval-100K}. Best in \textbf{bold}, second \underline{underlined}.}
    \vspace{-6pt}
   \resizebox{\linewidth}{!}{\begin{tabular}{l|cc:cc|cc:cc}
    \toprule
   \multirow{3}{40pt}{\textbf{Model (Visual Quality)}}  & \multicolumn{4}{c|}{\textbf{Q-Eval-Image}} & \multicolumn{4}{c}{\textbf{Q-Eval-Video}}  \\ \cdashline{2-9}
  &\multicolumn{2}{c}{\textit{Instance-level}}&\multicolumn{2}{c|}{\textit{Model-level}}&\multicolumn{2}{c}{\textit{Instance-level}}&\multicolumn{2}{c}{\textit{Model-level}} \\ \cdashline{2-9}
  &SRCC&PLCC&SRCC&PLCC&SRCC&PLCC&SRCC&PLCC \\ \hline
  NIQE & 0.239 & 0.238 & 0.829 & 0.758 & -0.057 & -0.074 & -0.333 & -0.241 \\
  CLIP-IQA & 0.334 & 0.324 & 0.600 & 0.685 & 0.194 & 0.175 & 0.095 & 0.076 \\
  Q-Align & \underline{0.587} & \underline{0.578} & 0.714 & 0.914 & \underline{0.500} & \underline{0.502} & \textbf{0.762} & \underline{0.762} \\
  IPCE & 0.550 & 0.560 & \underline{0.933} & \underline{0.937} & 0.299 & 0.302 & 0.476 & 0.568\\
  \textbf{Q-Eval-Score} & \textbf{0.732} & \textbf{0.731} & \textbf{0.943} & \textbf{0.949} & \textbf{0.601} & \textbf{0.609}  & \textbf{0.762} & \textbf{0.814} \\
    \bottomrule
\end{tabular}}
\vspace{-12pt}
    \label{tab:quality}
\end{table}

\begin{table}[!t]\small    
    \centering
    \renewcommand\arraystretch{1.08}
    \setlength{\tabcolsep}{4pt}
    \caption{Performance comparison on the alignment aspect of \textbf{Q-Eval-100K}. Considering that CLIPScore, BLIP2Score, and VQAScore are popular zero-shot alignment evaluation metrics, we provide the corresponding performance with the official default weight as well (marked with *).}
    \vspace{-6pt}
   \resizebox{\linewidth}{!}{\begin{tabular}{l|cc:cc|cc:cc}
    \toprule
   \multirow{3}{40pt}{\textbf{Model (Alignment)}}  & \multicolumn{4}{c|}{\textbf{Q-Eval-Image}} & \multicolumn{4}{c}{\textbf{Q-Eval-Video}}  \\ \cdashline{2-9}
  &\multicolumn{2}{c}{\textit{Instance-level}}&\multicolumn{2}{c|}{\textit{Model-level}}&\multicolumn{2}{c}{\textit{Instance-level}}&\multicolumn{2}{c}{\textit{Model-level}} \\ \cdashline{2-9}
  &SRCC&PLCC&SRCC&PLCC&SRCC&PLCC&SRCC&PLCC \\ \hline
  CLIPScore* & 0.245 & 0.252 & 0.617 & 0.685 & 0.186 & 0.219 & 0.518 & 0.500 \\
  BLIP2Score* & 0.297 & 0.330 & 0.764 & 0.835 & 0.218 & 0.250 & 0.295 & 0.296 \\
  VQAScore*  & 0.549 & 0.468 & 0.385 & 0.555 & 0.433 & 0.432 & 0.433 & 0.351 \\ \hdashline
  CLIPScore& \underline{0.768} & 0.740 & \underline{0.958} & \underline{0.956} & 0.431 & 0.443 & \underline{0.519} & \underline{0.509} \\
  BLIP2Score & 0.766 & \underline{0.743} & 0.933 & 0.934 & \underline{0.483} & \underline{0.488} & 0.512 & 0.481 \\
   ImageReward & 0.762 & 0.732 & 0.925 & 0.955 & 0.472 & 0.485 & 0.375 & 0.362 \\
  \textbf{Q-Eval-Score} & \textbf{0.822} & \textbf{0.802} & \textbf{0.964} & \textbf{0.969} & \textbf{0.607} & \textbf{0.634} & \textbf{0.648} & \textbf{0.605} \\
    \bottomrule
\end{tabular}}
\vspace{-12pt}
    \label{tab:alignment}
\end{table}

\section{Experiment}

\subsection{Experimental Setup}

\textbf{Training \& Evaluation.}
The Qwen2-VL-7B-Instruct~\cite{Qwen-VL} serves as the backbone LMM for \textbf{Q-Eval-Score}. All visual quality and alignment data from images and videos are combined for training. Training is conducted on 8 A100 GPUs for one epoch by default. 
For evaluation metrics, we use SRCC and PLCC, which measure the rank and linear correlation between predicted scores and MOSs. 
We propose evaluations at the \textbf{Instance-level} and \textbf{Model-level} which assess accuracy in ranking specific generated instances and generative models based on overall performance.


\begin{table*}[!t]\small    
    \centering
    \renewcommand\arraystretch{.95}
    \setlength{\tabcolsep}{6pt}
    \caption{Ablation Study of \textbf{Q-Eval-Score}.}
    \vspace{-6pt}
   \resizebox{\linewidth}{!}{\begin{tabular}{l|cc:cc:cc:cc|cc:cc:cc:cc}
    \toprule
   \multirow{3}{40pt}{\textbf{Model}}  & \multicolumn{4}{c:}{\textbf{Q-Eval-Image (Quality)}} & \multicolumn{4}{c|}{\textbf{Q-Eval-Video (Quality)}}  & \multicolumn{4}{c:}{\textbf{Q-Eval-Image (Alignment)}} & \multicolumn{4}{c}{\textbf{Q-Eval-Video (Alignment)}} \\ \cdashline{2-17}
  &\multicolumn{2}{c}{\textit{Instance-level}}&\multicolumn{2}{c:}{\textit{Model-level}}&\multicolumn{2}{c}{\textit{Instance-level}}&\multicolumn{2}{c|}{\textit{Model-level}} &\multicolumn{2}{c}{\textit{Instance-level}}&\multicolumn{2}{c:}{\textit{Model-level}}&\multicolumn{2}{c}{\textit{Instance-level}}&\multicolumn{2}{c}{\textit{Model-level}}  \\ \cdashline{2-17}
  &SRCC&PLCC&SRCC&PLCC&SRCC&PLCC&SRCC&PLCC&SRCC&PLCC&SRCC&PLCC&SRCC&PLCC&SRCC&PLCC \\ \hline
  \textit{w/o} \textbf{SFT Training} & 0.071 & 0.096 & 0.257 & 0.136 & 0.018 & 0.008 & 0.262 & 0.314 & 0.529 & 0.423 & 0.560 & 0.705 & 0.464 & 0.437 & 0.567 & 0.478 \\
  \textit{w/o} \textbf{ Context Prompt} & 0.504 & 0.509 & 0.600 & 0.756 & \underline{0.598} & \underline{0.591} & 0.571 & 0.638 & \underline{0.805} & \underline{0.776} & \underline{0.960} & \underline{0.963} & 0.588 & 0.597 & 0.601 &\underline{0.602}\\
  \textit{w/o} \textbf{ CE Loss} & 0.652 & 0.622 & 0.932 & 0.910 & 0.247 & 0.249 & 0.071 & 0.239 & 0.804 & \underline{0.776} & 0.948 & 0.961 & \underline{0.604} & \underline{0.626} & \underline{0.642} & 0.593 \\
  \textit{w/o} \textbf{ MSE Loss} & \underline{0.665} & \underline{0.673} & \underline{0.933} & \underline{0.941} & {0.595} & {0.583} & \underline{0.690} & \underline{0.712} & 0.795 & 0.763 & 0.954 & 0.958 & 0.580 & 0.605 & 0.624 & 0.599\\
  \textbf{Q-Eval-Score} & \textbf{0.732} & \textbf{0.731} & \textbf{0.943} & \textbf{0.949} & \textbf{0.601} & \textbf{0.609}  & \textbf{0.762} & \textbf{0.814} & \textbf{0.822} & \textbf{0.802} & \textbf{0.964} & \textbf{0.969} & \textbf{0.607} & \textbf{0.634} & \textbf{0.648} & \textbf{0.605}\\
    \bottomrule
\end{tabular}}
\vspace{-15pt}
    \label{tab:ablation}
\end{table*}

\begin{table}[!t]\small    
    \centering
    \renewcommand\arraystretch{.95}
    \setlength{\tabcolsep}{16pt}
    \caption{Performance comparison on the alignment aspect of \textbf{Q-Eval-100K} on the \textbf{long prompt subset}, where \textit{w/o} \textbf{V2S} and \textit{w} \textbf{V2S} represents the proposed \textbf{Q-Eval-Score} model \textit{with} and \textit{without} the \textbf{Vague-to-Specific} strategy respectively. }
    \vspace{-6pt}
   \resizebox{\linewidth}{!}{\begin{tabular}{l|cc|cc}
    \toprule
   \multirow{3}{40pt}{\textbf{Model (Alignment)}}  & \multicolumn{2}{c|}{\textbf{Q-Eval-Image (Long)}} & \multicolumn{2}{c}{\textbf{Q-Eval-Video (Long)}}  \\ \cdashline{2-5}
  &\multicolumn{2}{c|}{\textit{Instance-level}}&\multicolumn{2}{c}{\textit{Instance-level}} \\ \cdashline{2-5}
  &SRCC&PLCC&SRCC&PLCC \\ \hline
  CLIPScore & 0.533 & 0.547 & 0.359 & 0.367  \\
  BLIP2Score & 0.620 & 0.636 & 0.392 & 0.395  \\
  VQAScore & 0.432 & 0.325 & 0.344 & 0.350 \\ \hdashline
  \textit{w/o} \textbf{V2S} & \underline{0.591} & \underline{0.599} & \underline{0.480} & \underline{0.470} \\
  \textit{w} \textbf{V2S} & \textbf{0.620} & \textbf{0.623} & \textbf{0.517} & \textbf{0.512} \\
    \bottomrule
\end{tabular}}
\vspace{-15pt}
    \label{tab:alignment_long}
\end{table}

\noindent \textbf{Competitors.}
Few models can simultaneously predict both visual quality and alignment. Thus we selected task-specific competitors for each sub-task: For visual quality, we include NIQE~\cite{niqe}, CLIP-IQA~\cite{clipiqa}, Q-Align~\cite{wu2024qalign}, and IPCE~\cite{IPCE} (the top method from the \textit{`NTIRE 2024 Quality Assessment of AI-Generated Content Challenge'~\cite{liu2024ntire}}). For alignment, we choose CLIPScore~\cite{hessel2021clipscore}, BLIP2Score~\cite{li2023blip2}, ImageReward~\cite{imagereward} and VQAScore~\cite{vqascore} as the competitors. 
All models are trained and tested using their default recommended parameters and the corresponding train-test sets of the \textbf{Q-Eval-100K} dataset unless specified.

\subsection{Discussion \& General Findings}

The general performance on the visual quality and alignment is exhibited in Table~\ref{tab:quality} and Table~\ref{tab:alignment}, from which we can draw several conclusions: 1) For visual quality, The proposed \textbf{Q-Eval-Score} outperforms all competitors, achieving the best performance overall. The decline in video performance is likely due to the 1fps frame sampling method, which causes a loss of temporal information, leading to inaccurate estimations. Despite this, at the instance-level, \textbf{Q-Eval-Score} still leads the second-best competitor (Q-Align) by 10\% on video instance-level SRCC. 2) For alignment, \textbf{Q-Eval-Score} also demonstrates a significant lead in alignment, outperforming competitors by 6\% in image instance-level SRCC and 12\% in video instance-level SRCC. Additionally, the substantial performance improvements seen in the trained competitors suggest that \textbf{Q-Eval-100K} provides valuable knowledge for alignment evaluation. 3) In comparison to alignment, \textbf{Q-Eval-Score}’s performance in visual quality is notably lower, indicating that predicting visual quality is more challenging. This is likely because alignment evaluation is more straightforward and objective, while visual quality perception is more complex and subjective, making it harder to assess. Overall, the proposed \textbf{Q-Eval-Score} exhibits exceptional potential in both visual quality and alignment, achieving over 0.94 performance at the image model-level, closely aligning with human evaluations. This strong performance not only highlights the robustness of the model but also underscores its promising ability to serve as an effective evaluation metric.

\begin{table}[!t]\small    
    \centering
    \renewcommand\arraystretch{.95}
    \setlength{\tabcolsep}{15pt}
    \caption{Cross-dataset validation performance on GenAI-Bench. The \textbf{Q-Eval-Score} is trained on the \textbf{Q-Eval-100K} and then validated on GenAI-Bench. * indicates using default weight.}
    \vspace{-6pt}
   \resizebox{\linewidth}{!}{\begin{tabular}{l|cc|cc}
    \toprule
   \multirow{2}{60pt}{\textbf{Model (Alignment)}}  & \multicolumn{2}{c|}{\textbf{GenAI-Bench (Image)}} & \multicolumn{2}{c}{\textbf{GenAI-Bench (Video)}}  \\ \cdashline{2-5}
  &SRCC&PLCC&SRCC&PLCC \\ \hline
  CLIPScore* & 0.174 & 0.169 & 0.269 & 0.269\\
  BLIP2Score* & 0.221 & 0.209 & 0.289 & 0.275 \\
  VQAScore* & 0.556 & 0.502 & 0.527 & 0.505 \\ \hdashline
  CLIPScore & 0.681 & 0.670 & 0.610 & 0.628   \\
  BLIP2Score & \underline{0.687} & \underline{0.679} & \underline{0.679} & \underline{0.705}   \\
  ImageReward & 0.664 & 0.656 & 0.663 & 0.684 \\
  
  \textbf{Q-Eval-Score} & \textbf{0.757}& \textbf{0.747} & \textbf{0.717} & \textbf{0.714} \\
    \bottomrule
\end{tabular}}
\vspace{-15pt}
    \label{tab:alignment_cross}
\end{table}

\subsection{Further Experiments}

\noindent \textbf{I) Ablation Study.}
We conduct a detailed ablation study to assess the contribution of proposed strategies and CE/MSE loss. The results are presented in Table~\ref{tab:ablation}. It is clear that each of the strategies we proposed and both CE/MSE loss make a significant contribution to the final outcome.

\noindent \textbf{II) Long Prompt.}
To test the \textbf{Vague-to-Specific} strategy for long prompt alignment, we select a subset of 5,000 instances from \textbf{Q-Eval-100K} that contain long prompts (over 25 words) for testing, performance shown in Table~\ref{tab:alignment_long}. Due to the limited data size, we present only the instance-level performance. The results clearly show that the \textbf{Vague-to-Specific} strategy significantly improves performance, indicating the effectiveness of handling long prompt alignment.

\noindent \textbf{III) Cross-dataset Validation.}
To demonstrate the value of the \textbf{Q-Eval-100K} dataset, we conduct a cross-dataset validation, with performance results shown in Table~\ref{tab:alignment_cross}. It is important to note that instances generated from prompts in GenAI-Bench are excluded from this validation. The results clearly show that models trained on \textbf{Q-Eval-100K} significantly outperform the current state-of-the-art VQAScore on GenAI-Bench by a large margin, providing strong evidence of the generalization value of the \textbf{Q-Eval-100K} dataset.

\section{Conclusion}
In conclusion, we introduce \textbf{Q-Eval-100K}, the largest text-to-vision evaluation dataset to date, featuring 100K instances and 960K human annotations for assessing visual quality and alignment. We also present \textbf{Q-Eval-Score}, a unified evaluation framework that leverages this dataset to provide separate scores for each dimension. Experimental results show that \textbf{Q-Eval-Score} outperforms existing methods, demonstrating its potential for more reliable, comprehensive assessments of text-to-vision models. Looking ahead, we hope this work can lay a strong foundation for further advancements in text-to-vision model promotion and real-world evaluation applications.

\section*{Acknowledgment}
The work was supported in part by the National Natural Science Foundation of China under Grant 62301310, Grant 623B2073, and in part by Sichuan Science and Technology Program under Grant 2024NSFSC1426.


{
    \small
    \bibliographystyle{ieeenat_fullname}
    \bibliography{main,append}
}

\clearpage
\setcounter{page}{1}
\maketitlesupplementary

\section{Dataset Construction Details}

In this section, we mainly talk about the details of prompts collection.

\subsection{Prompts Collection}

The prompt collection comprises two sources:
\begin{itemize}
    \item \textbf{Internally Constructed Prompts}, which is based on internal capability requirement of Q-Eval-100K.
    \item \textbf{Open-Source Prompts}, which is based on other text-to-vison alignment evaluation datasets, icluding GenAIbench~\cite{li2024genaibench} and Docci~\cite{onoe2024docci}. GenAIbench features comprehensive prompt designs, while Docci provides longer prompts, making it suitable for evaluating long-prompt descriptions.
\end{itemize}

\subsection{Internally Constructed Prompts}
\begin{itemize}
    \item \textbf{Manual Construction:} Data is manually created by searching for commonly used prompts and rewriting them to align with the distribution of specific capabilities.
    \item \textbf{GPT-4 Augmentation:} GPT-4 is used to expand the dataset for specific capabilities. This process involves leveraging a few manually constructed examples and applying a Chain-of-Thought (CoT) approach. GPT-4 generates prompts based on given definitions and examples, which are then filtered and refined manually.
\end{itemize}

\noindent \textbf{Example GPT-4 Prompt Generation Instruction:}

\noindent \textit{You are an expert at crafting text-to-image prompts. I need prompts for text-to-image models based on the following category labels. Each label is explained with a description, the text before the ';' is the label, and the text after the ';' provides details.
Use your imagination and creativity to generate relevant English prompts. 
The prompt length should be between [len\_min] and [len\_max]. Avoid extra content, only output prompts. }

\subsection{Prompt Designing}
As shown in Table~\ref{tab:prompt_design}, the prompt design focuses on three main aspects: \textbf{Entity Generation}, \textbf{Entity Attribute Generation}, and \textbf{Interaction Capability}. 1) Entity generation targets the primary entities (\textit{people, objects, etc.}) to be generated. 2) Entity attribute generation emphasizes the attributes (\textit{clothing, color, material, visual quality, etc.}) of the entities.  3) Interaction capability focuses on the interactions between the generated entities and other entities or the background, such as their spatial relationships and actions.

\begin{table}[t]\small
\centering
\renewcommand\arraystretch{1.02}
\setlength{\tabcolsep}{9pt}
\caption{Detailed Descriptions of Entity Generation, Entity Attribute Generation, and Interaction Ability}
\label{tab:classification}
\resizebox{\linewidth}{!}{\begin{tabular}{lll}
\toprule
\textbf{Category}                  & \textbf{Subcategory}                                   & \textbf{Count} \\ \midrule
\multirow{9}{*}{Entity Generation} 
                                   & Simple Entity Generation                               & 1439           \\ \cdashline{2-3}
                                   & \quad Simple Human Generation                  & /  \\
                                   & \quad Simple Object Generation                  & / \\
                                   & \quad Other Simple Entity Generation                  & /            \\ \cline{2-3}
                                   & Complex Entity Generation                              & 1729           \\
                                   & \quad Character Information Generation                      & /               \\
                                   & \quad Text and Symbol Generation                      & /            \\
                                   & \quad Chart Generation                                & /          
                                      \\ \midrule
\multirow{27}{*}{Entity Attribute Generation} 
                                   & Basic Entity Attributes                               &    1656            \\ \cdashline{2-3}
                                   & \quad Entity Shape Generation                         & /            \\
                                   & \quad Entity Position Generation                      & /            \\
                                   & \quad Entity Color Generation                         & /            \\
                                   & \quad Entity State Generation                         & /            \\
                                   & \quad Other Entity Attributes Generation              & /            \\ \cline{2-3}
                                   & Person and Animal Attributes Generation               & 1500           \\ \cdashline{2-3}
                                   & \quad Emotion Generation                              & /            \\
                                   & \quad Action Generation                               & /            \\
                                   & \quad Specific Age Person Generation                  & /            \\
                                   & \quad Specific Gender Person Generation               & /            \\
                                   & \quad Other Person and Animal Attributes              & /            \\ \cline{2-3}
                                   & Portrait Generation                                   &   531             \\ \cdashline{2-3}
                                   & \quad Simple Portrait Generation                      & /            \\
                                   & \quad Complex Portrait Generation                     & /           \\ \cline{2-3}
                                   & Scene and Theme Generation                            &    2450            \\ \cdashline{2-3}
                                   & \quad Theme Generation                                & /           \\
                                   & \quad Scene Generation                                & /           \\ \cline{2-3}
                                   & Style Generation                                      & 294            \\ \cline{2-3}
                                   & Basic Visual Attributes Generation                    &  321              \\ \cdashline{2-3}
                                   & \quad Image Sharpness Generation                      & /            \\
                                   & \quad Exposure Generation                             & /            \\
                                   & \quad Lighting Generation                             & /            \\
                                   & \quad Contrast Generation                             & /            \\
                                   & \quad Color Saturation Generation                     & /            \\
                                   & \quad Noise Level Generation                          & /            \\
                                   & \quad Composition Generation                          & /            \\
                                   & \quad Color Balance Generation                        & /            \\
                                   & \quad Depth of Field Generation                       & /            \\
                                   & \quad Perspective Generation                          & /            \\
                                   & \quad Camera Angle Generation                         & /            \\
                                   & \quad Other Basic Visual Attributes Generation        & /     
                                           \\ \midrule
\multirow{4}{*}{Interaction Ability} 
                                   & Interacting Multi-Entity Generation                   & 1729           \\ \cdashline{2-3}
                                   & \quad Sequential Relationship  Multi-Entity Generation        & /            \\
                                   & \quad Causal Relationship  Multi-Entity Generation            & / \\ 
                                   & \quad Spatial Relationship  Multi-Entity Generation            & /  \\ 
\bottomrule
\label{tab:prompt_design}
\end{tabular}}
\vspace{-12pt}
\end{table}

\section{Long Prompt Split}

We use Qwen-VL-72B-Instruct~\cite{Qwen-VL} to help summarize the long prompt and split the long prompt into short sentences. Specifically, the prompt is designed as follows:

\noindent \textbf{Summarize Prompt}

\noindent \textit{\# User:  Please shorten the prompt to between 15 and 25 words, retaining the main information and ignoring details, specifically the characters, attributes, actions, and scenes. The prompt is as follows [Prompt].}

\noindent \textbf{Split Prompt}

\noindent \textit{\# User:  Split the prompt into three or fewer shorter prompts, with each short prompt describing one aspect of the original long prompt's subject and should be fewer than 15 words. The prompt is as follows [Prompt].}

\begin{figure*}
    \centering
    \subfloat[Overall performance on \textbf{Visual Quality}.]{
        \includegraphics[width=.41\linewidth]{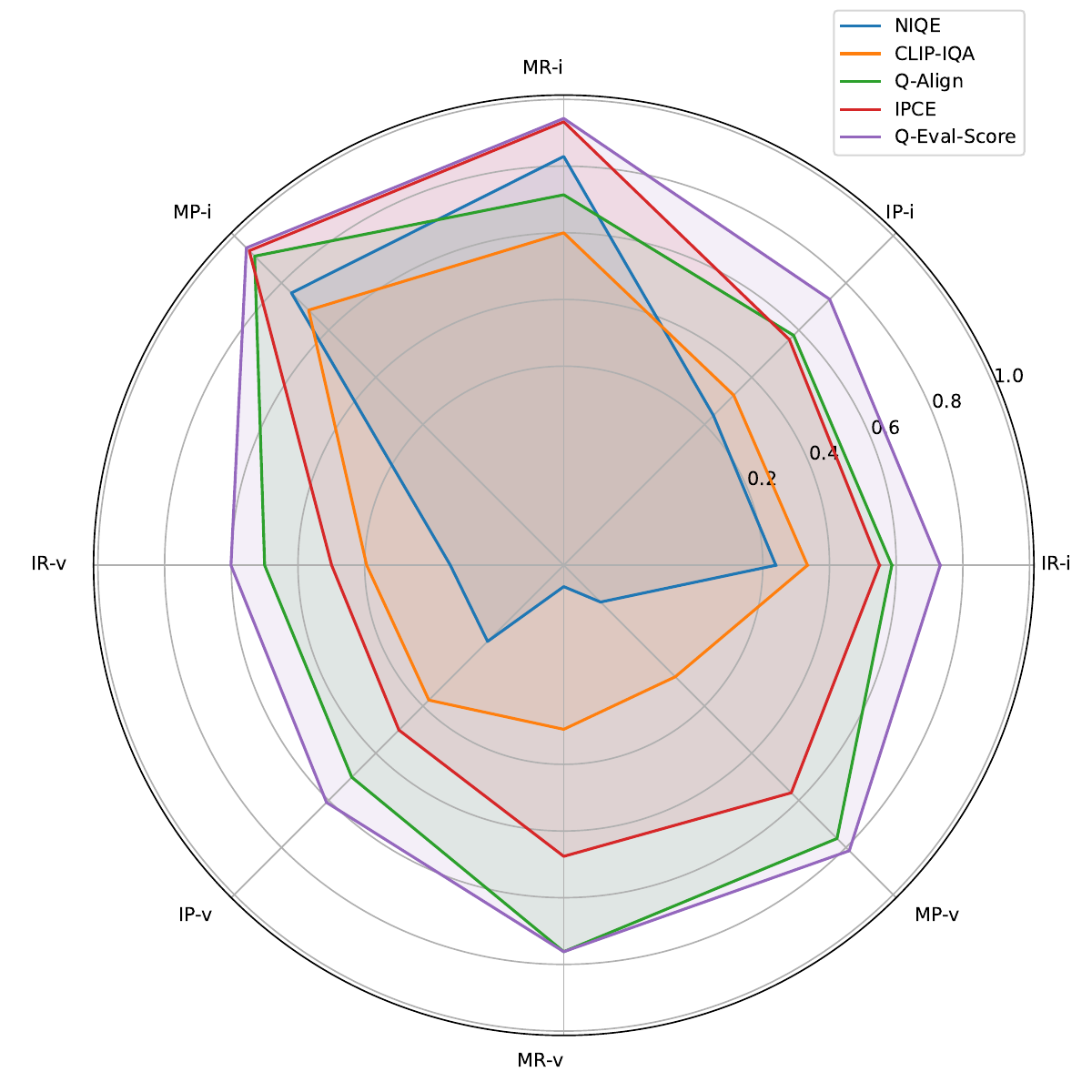}
        \label{fig:performance}
    }
    \hfill  
    \subfloat[Overall performance on \textbf{Alignment}.]{
        \includegraphics[width=.47\linewidth]{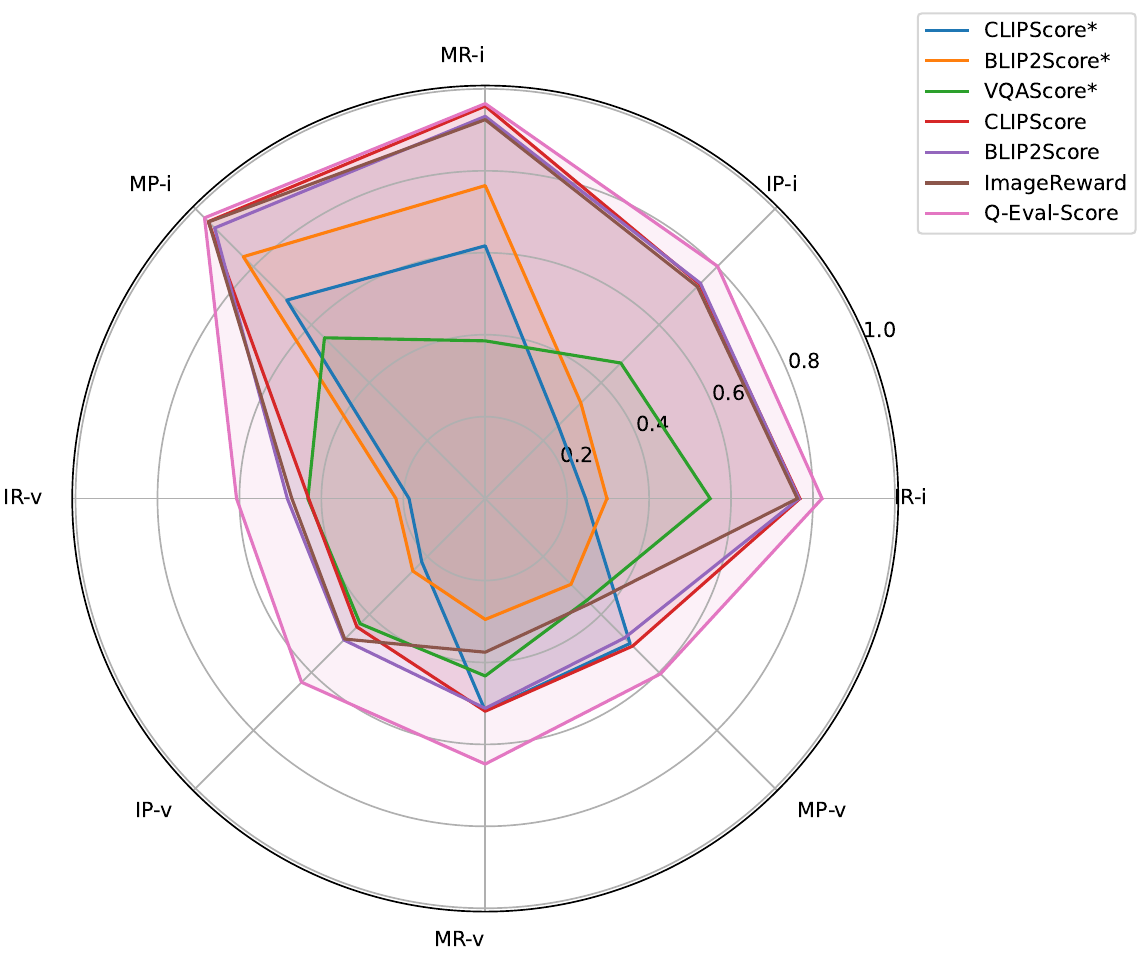}  
        \label{fig:table}
    }
    \caption{Radar charts of the overall performance on the \textbf{Visual Quality} and \textbf{Alignment} aspects on \textbf{Q-Eval-100K}, where IR, IP, MR, MP indicate Instance-level SRCC, Instance-level PLCC, Model-level SRCC, Model-level PLCC, and -i, -v represents image and video respectively.
 }
    \label{fig:radar}
\end{figure*}

\begin{figure*}
    \centering
    \includegraphics[width=\linewidth]{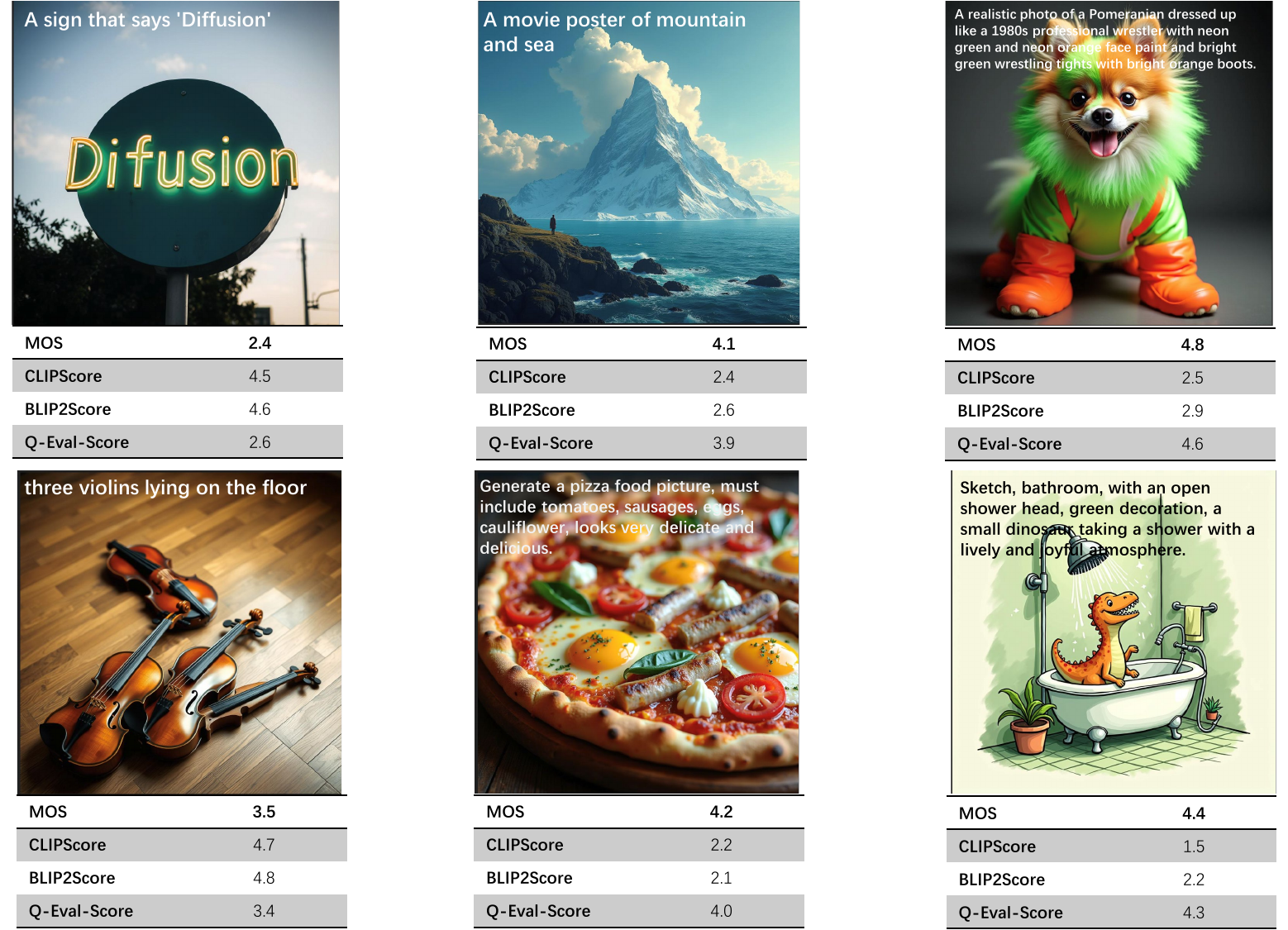}
    \caption{Visualization comparison results.}
    \label{fig:visual}
    \vspace{-6pt}
\end{figure*}

\subsection{ Subjective experiment details:}

\begin{figure}[t]\footnotesize
    \centering
    \begin{subfigure}{0.49\linewidth}
        \centering
        \includegraphics[width=\linewidth]{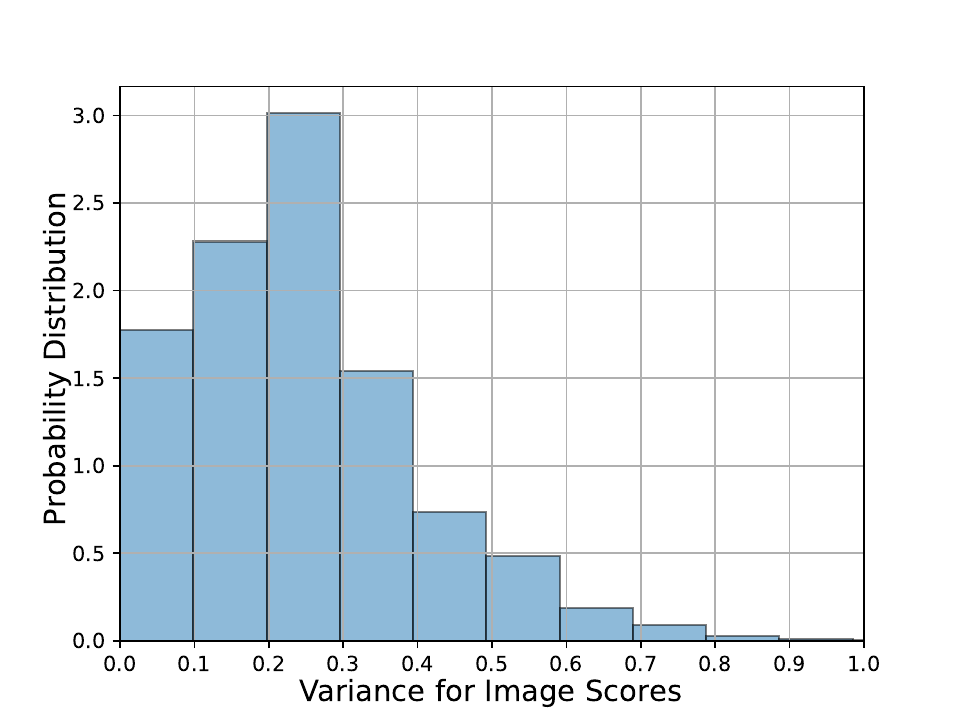}
        \label{fig:variance_img}
    \end{subfigure}
    \begin{subfigure}{0.49\linewidth}
        \centering
        \includegraphics[width=\linewidth]{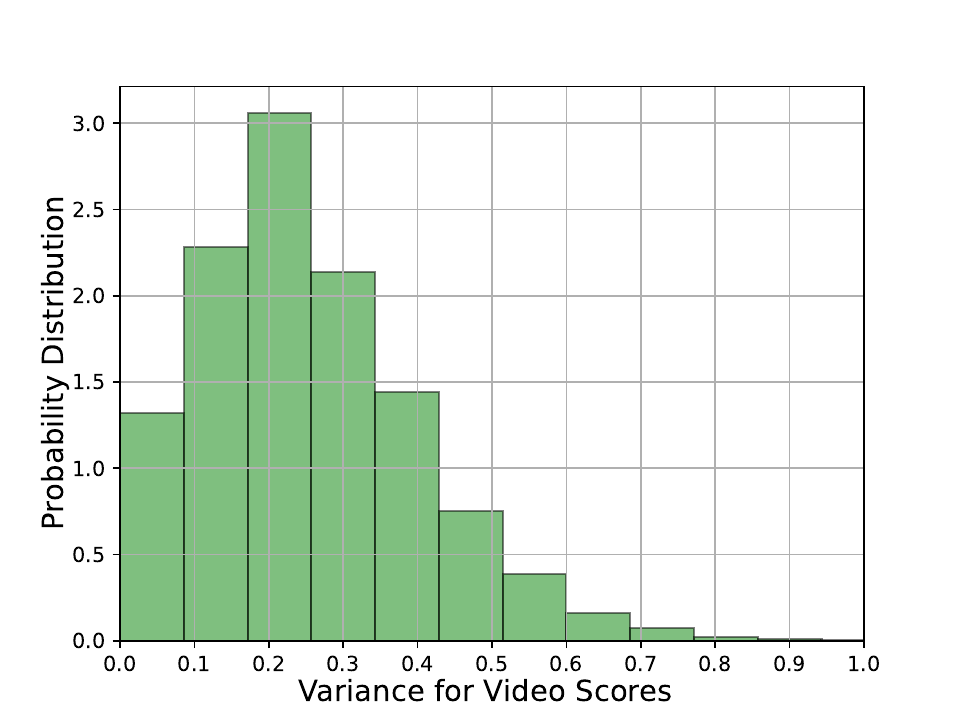}
        \label{fig:variance_video}
    \end{subfigure}
    \vspace{-12pt}
    \caption{Variance probability distributions for images/videos of Q-Eval-100K respectively. }
    \label{fig:var}
    \vspace{-15pt}
\end{figure}

1) Each instance in the training and test sets is rated by at least 3 and 12 individuals on average.
2) We ensure raters' diversity by employing raters from a wide age range (18-52) and selecting raters from various professional backgrounds.
Each rater annotates a maximum of 30 instances at a time, followed by a mandatory 30-minute break.
3) Perfect-score cases are rated by 12 individuals first, then reviewed and adjusted by a group of 5 experts.
4) Given the scale of Q-Eval-100K (\textit{the largest AIGC QA dataset with MOS at the cost of about 150,000 US dollars in total}), involving 15 raters per instance as suggested by ITU~\cite{itu} would be impractical due to time and cost constraints. To preserve the dataset's scale (\textit{crucial for LMM training under scaling laws}), we reduce the number of raters and implement a `\textit{Sample and Scrutinize}' approach to maintain annotation quality.
5) The variance distribution of instance annotations is shown in Fig.~\ref{fig:var}, where most instances have a variance below 0.3. 
\vspace{-8pt}

\section{Performance Details}

\subsection{Radar Charts of Overall Performance}
To provide a comprehensive overview of the performance, we present the radar charts in Fig.~\ref{fig:radar}. The key observations are as follows:
\begin{itemize}
    \item \textbf{Visual Quality.} The proposed \textbf{Q-Eval-Score} outperforms all competitors, achieving the highest overall performance. The slight decline in video performance can be attributed to the 1fps frame sampling method, which reduces temporal information and affects accuracy. Despite this limitation, \textbf{Q-Eval-Score} leads the second-best competitor (Q-Align) by a notable margin of 10\% in video instance-level SRCC.
    \item  \textbf{Alignment.}
    \textbf{Q-Eval-Score} also excels in alignment evaluation, surpassing competitors by 6\% in image instance-level SRCC and 12\% in video instance-level SRCC. Furthermore, the significant performance gains seen in other trained models indicate that \textbf{Q-Eval-100K} serves as a valuable dataset for improving alignment evaluation methods.
    \item \textbf{Comparison Between Tasks.}
    The performance of \textbf{Q-Eval-Score} in visual quality evaluation is relatively lower than in alignment tasks, highlighting the greater complexity of predicting visual quality. Alignment evaluation is more straightforward and objective, while visual quality involves nuanced and subjective perception, making it more challenging to assess.
\end{itemize}
Overall, the proposed \textbf{Q-Eval-Score} demonstrates remarkable capabilities in both visual quality and alignment evaluation. With performance exceeding 0.94 in image model-level metrics, it aligns closely with human judgment. These results underscore the robustness of \textbf{Q-Eval-Score} and its potential as a highly effective evaluation metric.

\subsection{More Cross-validation Results} 

We further select 4 datasets for cross-validation (See Table~\ref{tab:2}). AGIQA$^{Quality}$~\cite{agiqa3k} \& T2VQA~\cite{kou2024subjective} are for visual quality, while AGIQA$^{Align}$~\cite{agiqa3k} \&  TIFA160~\cite{hu2023tifa} are for text alignment. The results show good generalization ability of Q-Eval-Score. (best in \textbf{bold})

\begin{table}[h]\footnotesize
\centering
\caption{Cross-validation (All pre-trained on Q-Eval-100K).}
\vspace{-0.22cm}
\setlength{\tabcolsep}{4pt}
\renewcommand\arraystretch{1}
\resizebox{\linewidth}{!}{\begin{tabular}{c|cc|cc}
\toprule
\textbf{Dataset} & \textbf{AGIQA$^{Quality}$} & \textbf{T2VQA} & \textbf{AGIQA$^{Align}$} & \textbf{TIFA160}\\ \hline
Method & \textit{SRCC/PLCC} & \textit{SRCC/PLCC} & \textit{SRCC/PLCC} & \textit{SRCC/PLCC} \\ \hline
Q-Align & 0.6581/0.6743 & 0.2539/0.2198 & \textit{Inapplicable} & \textit{Inapplicable}\\
CLIPScore & \textit{Inapplicable} & \textit{Inapplicable} &0.5515/0.5627 &0.5903/0.5952 \\
BLIP2Score & \textit{Inapplicable} & \textit{Inapplicable} &0.6873/0.7085 &0.7267/0.7468 \\
\textbf{Q-Eval-Score} & \textbf{0.7256}/\textbf{0.7248} & \textbf{0.4479}/\textbf{0.4498} & \textbf{0.7544}/\textbf{0.7432} & \textbf{0.7845}/\textbf{0.7954}\\
\bottomrule
\end{tabular}}
\label{tab:2}
\end{table}

\subsection{Visualization Results}
We provide additional comparison examples in Fig.~\ref{fig:visual} to offer a clearer understanding of the evaluation capabilities of different models. It is evident from these examples that both CLIPScore and BLIPScore struggle significantly in tasks such as recognizing text within images and accurately counting objects. These models often fail by assigning disproportionately high scores to results that do not align with the intended outputs, reflecting their limitations in fine-grained assessment. Furthermore, when dealing with complex scenarios involving long and detailed prompts, these models exhibit a consistent tendency to assign significantly lower alignment scores, likely due to their inability to effectively parse and match intricate contextual information.
In contrast, \textbf{Q-Eval-Score} consistently demonstrates a much higher degree of accuracy and reliability in these challenging scenarios.  These results further highlight the potential of \textbf{Q-Eval-Score} as a unified framework for evaluating text-to-vision generative models across diverse and demanding conditions.

\section{Data Statement}
Considering the large scale of the dataset and the complexity of the model, we are actively organizing and refining the content to ensure its quality and usability. We solemnly pledge to release the \textbf{Q-Eval-100K} dataset in carefully planned batches, ensuring a comprehensive and systematic open-sourcing process that effectively supports community development. \textbf{Furthermore, we confirm that the dataset has successfully passed ethical review, affirming our commitment to responsible AI practices.} Alongside the dataset, we will also release \textbf{Q-Eval-Score} and provide continuous updates, ensuring the model remains aligned with the rapid advancements in generative AI.

\section{Broader Impact and Limitations}

\subsection{Broader Impact}
\begin{itemize}
    \item \textbf{Empowering Generative AI Applications.} 
     The development of comprehensive evaluation methods, such as Q-Eval-100K and Q-Eval-Score, directly supports these advancements by ensuring the quality and alignment of generated content, enabling its effective deployment.
    \item \textbf{Driving Standardization in Evaluation.}
    By introducing a unified framework for assessing visual quality and alignment, this work provides a benchmark for systematic evaluation. This standardization not only enhances the reliability of evaluations across diverse use cases but also fosters transparency in generative AI systems.
    \item \textbf{Facilitating Improvements in Generative Models.}
    The dataset and framework encourage the refinement of generative models by providing detailed feedback on visual quality and alignment. These insights guide iterative improvements, pushing the boundaries of what generative AI can achieve.
\end{itemize}

\subsection{Limitations}
\begin{itemize}
    \item \textbf{Subjectivity in Visual Quality Evaluation.}
    While Q-Eval-Score aligns closely with human evaluations, the inherently subjective nature of visual quality perception may result in variability. Differences in individual preferences and cultural factors could affect the generalizability of the evaluation framework.
    \item \textbf{Dependency on Human Annotations.}
    The reliance on extensive human annotations for dataset creation introduces scalability issues and potential biases. Automating parts of this process without sacrificing quality remains an open challenge.
\end{itemize}


\end{document}

%% file: preamble.tex
\usepackage{lineno}